\documentclass[twocolumn,showpacs,preprintnumbers,amsmath,amssymb]{revtex4}

\usepackage{graphicx}
\usepackage{dcolumn}
\usepackage{bm}

\newcommand{\8}{\infty}
\newcommand{\el}{\ell}

\newcommand{\be}{\begin{eqnarray*}}
\newcommand{\ee}{\end{eqnarray*}}
\newcommand{\beq}{\begin{equation}}
\newcommand{\eeq}{\end{equation}}
\newcommand{\beqn}{\begin{equation*}}
\newcommand{\eeqn}{\end{equation*}}
\newcommand{\bs}{\begin{split}}
\newcommand{\es}{\end{split}}

\begin{document}


\title{Physical models realizing the transformer architecture of large language models}

\author{Zeqian Chen}
\email{chenzeqian@hotmail.com}
\affiliation{%
Wuhan Institute of Physics and Mathematics, IAPM, Chinese Academy of
Sciences, West District 30, Xiao-Hong-Shan, Wuhan 430071, China}%


\begin{abstract}
The introduction of the transformer architecture in 2017 marked the most striking advancement in natural language processing. The transformer is a model architecture relying entirely on an attention mechanism to draw global dependencies between input and output. However, we believe there is a gap in our theoretical understanding of what the transformer is, and how it works physically. From a physical perspective on modern chips, such as those chips under 28nm, modern intelligent machines should be regarded as open quantum systems beyond conventional statistical systems. Thereby, in this paper, we construct physical models realizing large language models based on a transformer architecture as open quantum systems in the Fock space over the Hilbert space of tokens. Our physical models underlie the transformer architecture for large language models.
\end{abstract}


\maketitle


\section{Introduction}

Large language models (LLMs for short) are based on deep neural networks (DNNs) (cf.\cite{GBC2016}), and a common characteristic of DNNs is their compositional nature: data is processed sequentially, layer by layer, resulting in a discrete-time dynamical system. The introduction of the transformer architecture in 2017 marked the most striking advancement in terms of DNNs (cf.\cite{VSP2017}). Indeed, the transformer is a model architecture eschewing recurrence and instead relying entirely on an attention mechanism to draw global dependencies between input and output. At each step, the model is auto-regressive, consuming the previously generated symbols as additional input when generating the next. The transformer has achieved great success in natural language processing (cf.\cite{Zhao2023} and references therein).

The transformer has a modularization framework and is constructed by two main building blocks: self-attention and feed-forward neutral networks. Self-attention is an attention mechanism relating different positions of a single sequence in order to compute a representation of the sequence (cf.\cite{BCB2014}). In line with successful LLMs, one often focuses on the decoder-only setting of the transformer, where the model iteratively predicts the next tokens based on a given sequence of tokens. This procedure is coined autoregressive since the prediction of new tokens is only based on previous tokens. Such conditional sequence generation using autoregressive transformers is referred to as the transformer architecture. However, despite its meteoric rise within deep learning, we believe there is a gap in our theoretical understanding of what the transformer is, and how it works physically (cf.\cite{Mina2024}).

To the best of our knowledge, physical models for the transformer architecture of LLMs are usually described by using systems of mean-field interacting particles (cf. \cite{GLPR2025, VBC2020} and references therein), i.e., LLMs are regarded as classical statistical systems. However, from a physical perspective on modern chips, such as those chips under 28nm, modern intelligent machines process data through controlling the flow of electric current, i.e., the dynamics of largely many electrons, thereby, as indicated by Wilczek \cite{Wilczek2025}, LLMs should be regarded as open quantum systems beyond conventional statistical systems (cf.\cite{BP2002, VBR2025}). In this paper, we construct physical models realizing LLMs based on a transformer architecture as open quantum systems in the Fock space over the Hilbert space of tokens. Our physical models shed light on the fundamental behaviors of modern LLMs.

\section{Transformer architecture}

In the transformer architecture of a large language model $\mathfrak{S},$ we assume the finite set $\mathbf{T}$ of tokens in $\mathfrak{S}$ has been embedded in $\mathbb{R}^d,$ where $d$ is called the embedding dimension, so we identify each $t \in \mathbf{T}$ with one of finitely-many vectors $x$ in $\mathbb{R}^d.$ We assume that the structure (positional information, adjacency information, etc) is encoded in these vectors.  A finite sequence $\{x_i\}^n_{i=1}$ of tokens is called a text for $\mathfrak{S},$ simply denoted by $T= x_1 x_2 \cdots x_n$ or $(x_1, x_2, \cdots, x_n),$ where $n$ is called the length of the text $T.$ We write $[n] = \{1,2, \ldots,n\}$ for an integer $n.$

Recall that a self-attention layer {\rm SelfAtt} with an attention mechanism $(W^Q, W^K, W^V)$ in the transformer architecture is the only layer that combines different tokens, where $W^Q$ and $W^K$ are two $d' \times d$ real matrixes (i.e., the query and key matrixes) and $W^V$ is the $d \times d$ real matrix (called the value matrix) such that $W^V x \in \mathbf{T}$ for $x \in \mathbf{T}.$ Let us denote the input text to the layer by $X = \{x_i\}^n_{i=1}.$ For each $i \in [n],$ letting
\be
s_i = \frac{1}{\sqrt{d}} \langle W^Q x_n, W^Kx_i \rangle,\quad \forall i \in [n],
\ee
we can interpret $S^{(n)} =\{s_i \}^n_{i=1}$ as similarities between the $n$-th token $x_n$ (i.e., the query) and the other tokens (i.e., keys). The softmax layer is given by
\be
\mathrm{softmax}(S^{(n)})_i = \frac{e^{s_i}}{\sum^n_{j=1}e^{s_j}},\quad \forall i \in [n],
\ee
which can be interpreted as the probability for the $n$-th query to ``attend" to the $i$-th key. Then the self-attention layer $\mathrm{SelfAtt}$ is defined by
\beq\label{eq:SellAttAuto}
\mathrm{SelfAtt}(X)_n = \sum^n_{i=1} \mathrm{softmax} (S^{(n)})_i W^V x_i,
\eeq
indicating that the output $W^V x_i$ occurs with the probability $\mathrm{softmax} (S^{(n)})_i,$ which is often referred to as the values of the token $x_i.$ Note that $(W^Q, W^K, W^V)$ are trainable parameters in the transformer architecture.

In the same block, a feed-forward neural network $\mathrm{FFN}$ is then applied to $W^V x_i$'s such that $y_i = \mathrm{FFN}(W^V x_i)$ with the probability $\mathrm{softmax} (S^{(n)})_i$ for each $i \in [n],$ and so the output is $x_{n+1} = y_i = \mathrm{FFN}\circ \mathrm{SelfAtt} (\{x_i\}^n_{i=1})$ for some $i \in [n].$ One can then apply the same operations to the extended sequence $x_1 x_2\cdots x_n x_{n+1}$ in a next block, obtaining $x_{n+2} = \mathrm{FFN}'\circ \mathrm{SelfAtt}' (\{x_i\}^{n+1}_{i=1}),$ to iteratively compute further tokens (there is usually a stopping criterion based on a special token).

Typically, a transformer of depth $L$ is defined by a composition of $L$ blocks, denoted by ${\rm Transf}_L,$ consisting of $L$ self-attention maps $\{{\rm SelfAtt}_\el \}^L_{\el=1}$ and $L$ feed-forward neural networks $\{{\rm FFN}_\el \}^L_{\el =1},$ i.e.,
\beq\label{eq:Transf}
{\rm Transf}_L = ({\rm FFN}_L \circ {\rm SelfAtt}_L) \circ \cdots \circ ({\rm FFN}_1 \circ {\rm SelfAtt}_1),
\eeq
where the indices of the layers {\rm SelfAtt} and {\rm FFN} in \eqref{eq:Transf} indicate the use of different trainable parameters in each of the block. Then, given an input text $T=x_1 \cdots x_n,$ ${\rm Transf}_L$ generates a text $y_{i_1} \cdots y_{i_L}$ with the joint probability
\beq\label{eq:TransfProb}\begin{split}
P_T (y_{i_1}, & \cdots, y_{i_L})\\
= & \mathrm{softmax} (S^{(n)}_1)_{i_1} \cdots \mathrm{softmax} (S^{(n+L-1)}_L)_{i_L},
\end{split}\eeq
where $\mathrm{softmax} (S^{(n+\el -1)_\el})_{i_\el}$ is given by the attention mechanism $(W^Q_\el, W^K_\el, W^V_\el)$ in the $\el$-th building block for each $\el =1, \ldots, L.$ We refer to \cite{ZhangEt2025} and references therein for the details of the transformer architecture and LLMs.

\section{Physical models for LLMs}

 For a Hilbert space $\mathbb{K},$ we use $\mathcal{L} (\mathbb{K})$ and $\mathcal{S}(\mathbb{K})$ respectively to denote the set of all linear bounded operators and the set of all density operators in $\mathbb{K}.$ In what follows, we shall construct a physical model realizing the joint probability distributions \eqref{eq:TransfProb} given by the transformer architecture \eqref{eq:Transf} for LLMs. To this end, consider a large language model $\mathfrak{S}$ with the set $\mathbf{T}$ of $N$ tokens embedded in $\mathbb{R}^d.$ Let $\mathbf{h}$ be the Hilbert space with $\{|x\rangle:\; x \in \mathbf{T}\}$ being an orthogonal basis, and we identity $x = |x\rangle$ for $x \in \mathbf{T}.$ Let $\mathbb{H}=\mathcal{F} (\mathbf{h})$ be the Fock space over $\mathbf{h},$ that is,
\be
\mathcal{F} (\mathbf{h}) = \mathbb{C} \oplus \bigoplus^\8_{n =1} \mathbf{h}^{\otimes n},
\ee
where $\mathbf{h}^{\otimes n}$ is the $n$-fold tensor product of $\mathbf{h}$ (cf.\cite{RS1980I}) In the sequel, for the sake of convenience, we involve the finite Fock space
\be
\mathbb{H}= \mathcal{F}^{(M)} (\mathbf{h}) = \mathbb{C} \oplus \bigoplus^M_{n =1} \mathbf{h}^{\otimes n}
\ee
where $M$ is an integer such that $M \gg N.$ Note that an operator $A^{(n)} = A_1 \otimes \cdots \otimes A_n \in \mathcal{L} (\mathbf{h}^{\otimes n})$ for $A_j \in \mathcal{L} (\mathbf{h})$ satisfies that for all $h^{(n)}= h_1\otimes \ldots \otimes h_n \in \mathbf{h}^{\otimes n},$
\be
A h^{(n)} = (A_1 h_1) \otimes \ldots \otimes (A_n h_n)  \in \mathbf{h}^{\otimes n},
\ee
and in particular, if $\rho_i \in \mathcal{S}(\mathbf{h})$ for $i \in [n],$ then $\rho^{(n)} = \rho_1 \otimes \cdots \otimes \rho_n \in \mathcal{S} (\mathbf{h}^{\otimes n}).$ Given $\alpha \in \mathbb{C}$ and a sequence $A^{(n)} \in \mathcal{L} (\mathbf{h}^{\otimes n})$ for $n \in [M],$ the operator $\mathrm{diag} (\alpha, A^{(1)},\cdots, A^{(M)}) \in \mathcal{L} (\mathbb{H})$ is defined by
\be
\mathrm{diag} (\alpha, A^{(1)},\cdots, A^{(M)}) \mathrm{h}^{(M)} = (\alpha c, A^{(1)} h^{(1)}, \cdots , A^{(M)}h^{(M)})
\ee
for every $\mathrm{h}^{(M)} = (c, h^{(1)}, \cdots, h^{(M)}) \in \mathbb{H}.$ In particular, if $\rho^{(n)} \in \mathcal{S} (\mathbf{h}^{\otimes n}),$ then
\be
\rho^{(M)} = \mathrm{diag}(0, 0^{(1)},\cdots, 0^{(n-1)}, \rho^{(n)}, 0^{(n+1)},\cdots, 0^{(M)}) \in \mathcal{S} (\mathbb{H}),
\ee
where $0^{(i)}$ denotes the zero operator in $\mathbf{h}^{\otimes i}$ for $i\ge 1.$

It suffices to construct a physical model in the Fock space $\mathbb{H}=\mathcal{F}^{(M)} (\mathbf{h})$ ($M \gg L$) for a transformer $\mathrm{Transf}_L$ \eqref{eq:Transf} with a composition of $L$ blocks, consisting of $L$ self-attention maps $\{{\rm SelfAtt}_\el=(W^Q_\el, W^K_\el, W^V_\el) \}^L_{\el=1}$ and $L$ feed-forward neural networks $\{{\rm FFN}_\el \}^L_{\el =1}.$ \begin{widetext}Precisely, let us denote the input text to the layer by $T = \{x_i\}^n_{i=1}.$ As noted above, for $\el =1, \ldots, L,$ one has
\beq
\mathrm{FFN}_\el \circ \mathrm{SelfAtt}_\el (T) = \sum^{n+\el -1}_{i=1} \mathrm{softmax}(S^{(n+\el -1)}_\el)_i \mathrm{FFN}_\el (W^V_\el x_i),
\eeq
where $S^{(n+\el -1)}_\el = \{s^{(\el)}_i\}^{n+\el -1}_{i=1}$ and
\beq
s^{(\el)}_i = \frac{1}{\sqrt{d}} \langle W^Q_\el x_{n+\el -1}, W^K_\el x_i \rangle,\quad \forall i \in [n+\el -1].
\eeq
\end{widetext}
A physical model needed to construct for $\mathrm{Transf}_L$ consist of an input $\rho (t_0)$ and a sequence of quantum operations $\{\mathcal{E}(t_\el, t_0)\}^L_{\el =1}$ in the Fock space $\mathbb{H}$ (cf.\cite{NC2001}), where $t_0 < t_1 < \cdots < t_L.$ We show how to construct this model step by step as follows.

To this end, we denote by $\Omega = \{\diamond\} \cup \mathbf{T}$ and by $2^\Omega$ the set of all subsets of $\Omega,$ and write $\mathbf{D} = (\{\omega\}: \omega \in \Omega).$ At first, for an input text $T=x_1 \cdots x_n,$ the input state $\rho_T$ is given as
\begin{widetext}
\beq
\rho_T = \rho (t_0) = \mathrm{diag}(0, 0^{(1)},\cdots, 0^{(n-1)}, |x_1\rangle\langle x_1| \otimes \ldots \otimes |x_n \rangle\langle x_n|, 0^{(n+1)}, \cdots ) \in \mathcal{S} (\mathbb{H}).
\eeq
Then there is a quantum operation $\mathcal{E}(t_1, t_0)$ in $\mathbb{H},$ depending only on the attention mechanism $(W^Q_1, W^K_1, W^V_1)$ and $\mathrm{FFN}_1,$ such that (see Appendix for the details)
\beq\label{eq:KopLLM-1}\begin{split}
\mathcal{E}(t_1, t_0) \rho (t_0)
= \sum^n_{i=1} \mathrm{softmax}(S^{(n)}_1)_i \mathrm{diag} (0, 0^{(1)}\cdots, 0^{(n)}, |x_1\rangle\langle x_1| \otimes \ldots \otimes |x_n \rangle\langle x_n| \otimes |y^{(1)}_i \rangle \langle y^{(1)}_i|,0^{(n+2)}, \cdots),
\end{split}\eeq
where $y^{(1)}_i=\mathrm{FFN}_1 (W^{V_1} x_i)$ and $\{y^{(1)}_i\}^n_{i=1} \subset \{|x\rangle:\; x \in \mathbf{T}\}.$ Define $X_1: 2^\Omega \mapsto \mathcal{E}(\mathbb{H})$ by
\beq
X_1 (\{\diamond\}) = \mathrm{diag}(1, I_\mathbf{h}, \ldots, I_{\mathbf{h}}^{\otimes n}, 0^{(n+1)}, I_{\mathbf{h}^{\otimes (n+2)}}, \cdots ),
\eeq
and for every $x \in \mathbf{T},$
\beq
X_1 (\{x\}) = \mathrm{diag}(0, 0^{(1)}, \ldots, 0^{(n)}, \underbrace{I_\mathbf{h} \otimes \cdots \otimes I_\mathbf{h}}_n \otimes |x \rangle \langle x|, 0^{(n+2)}, \cdots ).
\eeq
Making a measurement $(X_1, \mathbf{D})$ at time $t_1,$ we obtain an output $y^{(1)}_i$ with probability $\mathrm{softmax}(S^{(n)}_1)_i$ and, according to the von Neumann-L\"{u}ders reduction postulate (cf.\cite{NC2001}), the appropriate density operator to use for any further calculation is
\beq\begin{split}
\rho_\mathrm{red} (t_1)_i = & \frac{E^{(1)}_i \rho (t_1) E^{(1)}_i}{\mathrm{Tr}[E^{(1)}_i \rho (t_1)]}\\
 = &\mathrm{diag} (0, 0^{(1)}\cdots, 0^{(n)}, |x_1\rangle\langle x_1| \otimes \ldots \otimes |x_n \rangle\langle x_n| \otimes |y^{(1)}_i \rangle \langle y^{(1)}_i|,0^{(n+2)}, \cdots),
\end{split}\eeq
for every $i \in [n],$ where $\rho (t_1) = \mathcal{E}(t_1, t_0) \rho (t_0),$ and
\beq
E^{(1)}_i = \mathrm{diag}(0, 0^{(1)}, \ldots, 0^{(n)},  \underbrace{I_\mathbf{h}\otimes \ldots \otimes I_\mathbf{h}}_n \otimes |y^{(1)}_i \rangle \langle y^{(1)}_i|, 0^{(n+2)}, \cdots ).
\eeq

Next, there is a quantum operation $\mathcal{E}(t_2, t_0)$ in $\mathbb{H},$ depending only on the attention mechanism $(W^Q_2, W^K_2, W^V_2)$ and $\mathrm{FFN}_2$ at time $t_2,$ such that (see Appendix again)
\beq\label{eq:KopLLM-2}
\begin{split}
\mathcal{E}(t_2, t_0) \rho_\mathrm{red} & (t_1)_{i_1} =  \sum^{n+1}_{i_2 =1} \mathrm{softmax}(S^{(n+1)}_2)_{i_2} \\
\times \mathrm{diag}(0, & 0^{(1)},\cdots, 0^{(n+1)}, |x_1\rangle\langle x_1| \otimes \ldots \otimes |x_n \rangle\langle x_n| \otimes |y^{(1)}_{i_1} \rangle \langle y^{(1)}_{i_1}| \otimes |y^{(2)}_{i_2} \rangle \langle y^{(2)}_{i_2}|, 0^{(n+3)}, \cdots),
\end{split}
\eeq
for $i_1 \in [n],$ where $y^{(2)}_{i_2} = \mathrm{FFN}_2 (W^{V_2} x_{i_2})$ (with $x_{n+1} = y^{(1)}_{i_1}$) and $\{y^{(2)}_i\}^{n+1}_{i=1} \subset \{|x\rangle:\; x \in \mathbf{T}\}.$ Define $X_2: 2^\Omega \mapsto \mathcal{E}(\mathbb{H})$ by
\beq
X_2 (\{\diamond\}) = \mathrm{diag}(1, I_\mathbf{h}, \ldots, I_{\mathbf{h}}^{\otimes (n+1)}, 0^{(n+2)}, I_{\mathbf{h}^{\otimes (n+3)}}, \cdots ),
\eeq
and for every $x \in \mathbf{T},$
\beq
X_2 (\{x\}) = \mathrm{diag}(0, 0^{(1)}, \ldots, 0^{(n+1)}, \underbrace{I_\mathbf{h} \otimes \cdots \otimes I_\mathbf{h}}_{n+1} \otimes |x \rangle \langle x|, 0^{(n+3)}, \cdots ).
\eeq
Making a measurement $(X_2, \mathbf{D})$ at time $t_2,$ we obtain an output $y^{(2)}_{i_2}$ with probability $\mathrm{softmax}(S^{(n+1)}_2)_{i_2}$ and the appropriate density operator to use for any further calculation is
\beq\begin{split}
\rho_\mathrm{red} (t_2)_{i_1,i_2} = & \frac{E^{(2)}_{i_2} \rho (t_2)_{i_1} E^{(2)}_{i_2}}{\mathrm{Tr}[E^{(2)}_{i_2}\rho (t_2)_{i_1}]}\\
= & \mathrm{diag}(0, 0^{(1)},\cdots, 0^{(n+1)}, |x_1\rangle\langle x_1| \otimes \ldots \otimes |x_n \rangle\langle x_n| \otimes |y^{(1)}_{i_1} \rangle \langle y^{(1)}_{i_1}| \otimes |y^{(2)}_{i_2} \rangle \langle y^{(2)}_{i_2}|, 0^{(n+3)}, \cdots),
\end{split}\eeq
for each $i_2 \in [n+1],$ where $\rho (t_2)_{i_1} = \mathcal{E}(t_2, t_0) \rho_\mathrm{red} (t_1)_{i_1}$ and
\beq
E^{(2)}_{i_2} = \mathrm{diag}(0, 0^{(1)}, \ldots, 0^{(n+1)}, \underbrace{I_\mathbf{h}\otimes \ldots \otimes I_\mathbf{h}}_{n+1} \otimes |y^{(2)}_{i_2} \rangle \langle y^{(2)}_{i_2}|, 0^{(n+3)}, \cdots ).
\eeq

Step by step, we obtain a physical model $\{\mathcal{E}(t_\el, t_0)\}^L_{\el =1}$ with the input state $\rho_T = \rho (t_0)$ as given an input text $T=x_1 \cdots x_n,$ such that a text $(y^{(1)}_{i_1}, y^{(2)}_{i_2}, \ldots, y^{(L)}_{i_L})$ is generated with the probability
\be\begin{split}
\mathrm{P}_T (y^{(1)}_{i_1}, & y^{(2)}_{i_2}, \ldots, y^{(L)}_{i_L})
 =  \mathrm{softmax}(S^{(n)}_1)_{i_1} \cdots \mathrm{softmax}(S^{(n+L-1)}_L)_{i_L},
\end{split}\ee
within the sequential measurement $(X_1,\mathbf{D}), \ldots, (X_L,\mathbf{D}).$ Thus, the physical model so constructed realizes the joint probability distributions \eqref{eq:TransfProb} given by the transformer architecture $\mathrm{Transf}_L$ in \eqref{eq:Transf}.\end{widetext}

A physical model for the transformer with a multi-headed attention (cf.\cite{VSP2017}) can be constructed in a similar way. Also, we can construct physical models for the transformer of more complex structure (cf.\cite{ZhangEt2025} and reference therein). We omit the details.

{\bf Example.}\; Let $\mathbf{T}=\{x_0, x_1\}$ be the set of two tokens embedded in $\mathbb{R}^2$ such that $x_0 =(1,0)$ and $x_1=(0,1).$ Then $\mathbf{h}= \mathbb{C}^2$ with the standard basis $|0 \rangle =|x_0\rangle$ and $|1 \rangle =|x_1\rangle.$ Let $\mathbb{H} = \mathcal{F}^{(6)}(\mathbb{C}^2).$ Assume an input text $T = (x_0, x_1, x_0).$ The input state $\rho_T$ is then given by
\begin{widetext}
\be
\rho_T = \rho (t_0) = \mathrm{diag} (0, 0^{(1)}, 0^{(2)}, |x_0 \rangle \langle x_0| \otimes |x_1 \rangle \langle x_1|\otimes |x_0 \rangle \langle x_0|, 0^{(4)}, 0^{(5)}, 0^{(6)}).
\ee

If $W^Q_1 = W^V_1= \mathrm{FFN}_1= I$ and $W^K_1= \sigma_x$ in $\mathbb{R}^2,$ an associated physical operation $\mathcal{E}(t_1, t_0)$ at time $t_1$ satisfies
\be\begin{split}
\mathcal{E}(t_1, t_0) \rho (t_0) = & \frac{2}{e +2} \mathrm{diag} (0, 0^{(1)}, 0^{(2)}, 0^{(3)},|x_0 \rangle \langle x_0| \otimes |x_1 \rangle \langle x_1|\otimes |x_0 \rangle \langle x_0| \otimes |x_0 \rangle \langle x_0 | ,0^{(5)}, 0^{(6)})\\
+ & \frac{e}{e +2} \mathrm{diag} (0, 0^{(1)}, 0^{(2)}, 0^{(3)},|x_0 \rangle \langle x_0| \otimes |x_1 \rangle \langle x_1|\otimes |x_0 \rangle \langle x_0| \otimes |x_1 \rangle \langle x_1 | ,0^{(5)}, 0^{(6)}).
\end{split}\ee
By measurement, we obtain $x_0$ with probability $\frac{2}{e +2}$ and obtain $x_1$ with probability $\frac{e}{e +2},$ while
\be\begin{split}
\rho_\mathrm{red} (t_1)_0 =& \mathrm{diag}(0, 0^{(1)}, 0^{(2)}, 0^{(3)},|x_0 \rangle \langle x_0| \otimes |x_1 \rangle \langle x_1|\otimes |x_0 \rangle \langle x_0| \otimes |x_0 \rangle \langle x_0 | ,0^{(5)}, 0^{(6)}),\\
\rho_\mathrm{red} (t_1)_1 =& \mathrm{diag}(0, 0^{(1)}, 0^{(2)}, 0^{(3)},|x_0 \rangle \langle x_0| \otimes |x_1 \rangle \langle x_1|\otimes |x_0 \rangle \langle x_0| \otimes |x_1 \rangle \langle x_1 | ,0^{(5)}, 0^{(6)}).
\end{split}\ee

If $W^Q_2 =W^K_2 = \mathrm{FFN}_2 =I$ and $W^V_2 = \sigma_x$ in $\mathbb{R}^2,$ an associated quantum operation $\mathcal{E}(t_2, t_0)$ at time $t_2$ satisfies
\be\begin{split}
\mathcal{E}(t_2,& t_0) \rho_\mathrm{red} (t_1)_0\\
=& \frac{1}{3 e +1} (0, 0^{(1)}, 0^{(2)}, 0^{(3)}, 0^{(4)},|x_0 \rangle \langle x_0| \otimes |x_1 \rangle \langle x_1|\otimes |x_0 \rangle \langle x_0| \otimes |x_0 \rangle \langle x_0 | \otimes |x_0 \rangle \langle x_0 |, 0^{(6)} )\\
& + \frac{3e}{3 e +1} (0, 0^{(1)}, 0^{(2)}, 0^{(3)}, 0^{(4)},|x_0 \rangle \langle x_0| \otimes |x_1 \rangle \langle x_1|\otimes |x_0 \rangle \langle x_0| \otimes |x_0 \rangle \langle x_0 | \otimes |x_1 \rangle \langle x_1 |, 0^{(6)} ).
\end{split}\ee
and
\be\begin{split}
\mathcal{E}(t_2,& t_0) \rho_\mathrm{red} (t_1)_1\\
=& \frac{e}{e +1}(0, 0^{(1)}, 0^{(2)}, 0^{(3)},0^{(4)}, |x_0 \rangle \langle x_0| \otimes |x_1 \rangle \langle x_1|\otimes |x_0 \rangle \langle x_0| \otimes |x_1 \rangle \langle x_1 | \otimes |x_0 \rangle \langle x_0|, 0^{(6)} )\\
& + \frac{1}{e +1}(0, 0^{(1)}, 0^{(2)}, 0^{(3)},0^{(4)}, |x_0 \rangle \langle x_0| \otimes |x_1 \rangle \langle x_1|\otimes |x_0 \rangle \langle x_0| \otimes |x_1 \rangle \langle x_1 | \otimes |x_1 \rangle \langle x_1|, 0^{(6)} ).
\end{split}\ee
\end{widetext}
By measurement at time $t_2,$ when $x_0$ occurs at $t_1,$ we obtain $x_0$ with probability $\frac{1}{3 e +1}$ and obtain $x_1$ with probability $\frac{3 e}{3 e +1};$ when $x_1$ occurs at $t_1,$ we obtain $x_0$ with probability $\frac{e}{e +1}$ and obtain $x_1$ with probability $\frac{1}{e +1}.$

Hence, we obtain the joint probability distributions:
\be\begin{split}
P_T (x_0, x_0) = & \frac{2}{e +2} \frac{1}{3 e +1} = \frac{2}{(e+2)(3 e+1)},\\
P_T (x_0,x_1) = & \frac{2}{e +2} \frac{3e}{3 e +1}= \frac{6 e}{(e+2)(3 e+1)},\\
P_T (x_1, x_0) = & \frac{e}{e +2} \frac{e}{e +1} = \frac{e^2}{(e+2)(e+1)},\\
P_T (x_1,x_1) = & \frac{e}{e +2} \frac{1}{e +1}= \frac{e}{(e+2)(e+1)}.
\end{split}\ee

\section{Conclusions}\label{Concl}

In conclusion, we construct physical models realizing LLMs based on a transformer architecture as open quantum systems in the Fock space over the Hilbert space of tokens. Note that physical models satisfying the joint probability distributions associated with a transformer $\mathrm{Transf}_L$ are not necessarily unique. However, a physical model $\{\mathcal{E}(t_\el, t_0)\}^L_{\el =1}$ uniquely determines the joint probability distributions, that is, it defines a unique physical process for operating the large language model based on $\mathrm{Transf}_L.$ Therefor, in a physical model $\{\mathcal{E}(t_\el, t_0)\}^L_{\el =1}$ for $\mathrm{Transf}_L,$ training for $\mathrm{Transf}_L$ corresponds to training for the quantum operations $\{\mathcal{E}(t_\el, t_0)\}^L_{\el=1},$ which are adjustable and learned during the training process, determining the physical model, as corresponding to the parameter matrixes $\{(W^Q_\el, W^K_\el, W^V_\el)\}^L_{\el=1}$ in $\mathrm{Transf}_L.$ From a physical perspective, training for a large language model is just to determine the quantum operations $\{\mathcal{E}(t_\el, t_0)\}^L_{\el=1}$ associated with the corresponding open quantum system (cf.\cite{SC2022}). This means that our physical models underlie the transformer architecture for LLMs. We refer to \cite{Chen2025} for the details of mathematical formalism and physical models of generative AI and to \cite{CDLY2025} for a mathematical foundation of general AI, including quantum AI.

\



\section{Appendix}\label{App}

In this appendix, we show that if a building block consists of an attention mechanism $(W^Q, W^K, W^V)$ and $\mathrm{FFN}$ in a transformer architecture, then there is a quantum operation $\mathcal{E}$ in $\mathbb{H},$ depending only on $(W^Q, W^K, W^V)$ and $\mathrm{FFN},$ such that given an input text $T = \{x_i\}^n_{i=1},$ for the input state
\begin{widetext}
\be
\rho_T = \mathrm{diag}(0, 0^{(1)},\cdots, 0^{(n-1)}, |x_1\rangle\langle x_1| \otimes \ldots \otimes |x_n \rangle\langle x_n|, 0^{(n+1)}, \cdots ) \in \mathcal{S} (\mathbb{H}),
\ee
the quantum operation $\mathcal{E}$ satisfies
\be\begin{split}
\mathcal{E}\rho_T
= \sum^n_{i=1} \mathrm{softmax}(S^{(n)})_i \mathrm{diag} (0, 0^{(1)}\cdots, 0^{(n)}, |x_1\rangle\langle x_1| \otimes \ldots \otimes |x_n \rangle\langle x_n| \otimes |y_i \rangle \langle y_i|,0^{(n+2)}, \cdots),
\end{split}\ee
where $y_i= \mathrm{FFN} (W^{V} x_i)$'s ($i\in [n]$) are given by $(W^Q, W^K, W^V)$ and $\mathrm{FFN}.$

To this end, we regard $1,$ $|x\rangle\langle x|,$ and $|x_1\rangle \langle x_1| \otimes \ldots \otimes |x_n \rangle\langle x_n|$ as elements in $\mathcal{L} (\mathbb{H})$ in a natural way, i.e.,

\be\begin{split}
1 & \simeq \mathrm{diag}(1,0^{(1)}, 0^{(2)},\cdots ),\\
|x\rangle\langle x|& \simeq \mathrm{diag}(0,|x\rangle\langle x|, 0^{(2)},,\cdots ),\\
|x_1\rangle \langle x_1| \otimes \ldots \otimes |x_n \rangle\langle x_n|& \simeq \mathrm{diag}(0,0^{(1)},\cdots, 0^{(n-1)},|x_1\rangle \langle x_1| \otimes \ldots \otimes |x_n \rangle\langle x_n|, 0^{(n+1)}, \cdots ),
\end{split}\ee
for $n \ge 1.$ We first define
\be
\Phi (1) = |x_0\rangle\langle x_0|
\ee
where $x_0 \in \mathbf{T}$ is a certain element. Secondly, define
\be
\Phi (|x\rangle\langle x|)= \mathrm{diag}(0,0^{(1)}, |x\rangle\langle x| \otimes | \mathrm{FFN}(W^V x) \rangle \langle \mathrm{FFN}(W^V x)|,0^{(3)},\cdots),\quad \forall x \in \mathbf{T},
\ee
and in general, for $n \in [L]$ define
\be\begin{split}
\Phi (|x_1\rangle \langle x_1| & \otimes \ldots \otimes |x_n \rangle\langle x_n|)\\
= \sum^n_{i=1} & \mathrm{softmax}(S^{(n)})_i \mathrm{diag} (0, 0^{(1)}\cdots, 0^{(n)}, |x_1\rangle\langle x_1| \otimes \ldots \otimes |x_n \rangle\langle x_n| \otimes |y_i \rangle \langle y_i|,0^{(n+2)}, \cdots)
\end{split}\ee
for any $x_i \in \mathbf{T}$ and $i \in [n].$ Let
\be
\mathbf{S}= \mathrm{span}\{1, |x_1\rangle \langle x_1| \otimes \ldots \otimes |x_\el \rangle\langle x_\el|:\; x_i \in \mathbf{T},i \in [\el];\; \el=1, \ldots, L\}.
\ee
Then $\Phi$ extends uniquely to a positive map $\mathcal{E}_\Phi$ from $\mathbf{S}$ into $\mathcal{L} (\mathbb{H}),$ that is,
\be\begin{split}
\mathcal{E}_\Phi & \Big ( a_0 + \sum_{n \ge 1} \sum_{x_1, \cdots,x_n \in \mathbf{T}} a_{x_1, \cdots,x_n } |x_1\rangle\langle x_1| \otimes \ldots \otimes |x_n \rangle\langle x_n| \Big )\\
& = a_0 |x_0\rangle\langle x_0| + \sum_{n \ge 1} \sum_{x_1, \cdots,x_n \in \mathbf{T}} a_{x_1, \cdots,x_n } \Phi (|x_1\rangle\langle x_1| \otimes \ldots \otimes |x_n \rangle\langle x_n|),
\end{split}\ee\end{widetext}
where $a_0, a_{x_1, \cdots,x_n}$ are any complex numbers for $n \ge 1.$ Note that $\mathbf{S}$ is a commutative $C^*$-algebra. By Stinespring's theorem (cf.\cite[Theorem 3.11]{Paul2002}), $\mathcal{E}_\Phi: \mathbf{S} \mapsto \mathcal{L} (\mathbb{H})$ is completely positive. Hence, by Arveson's extension theorem (cf.\cite[Theorem 7.5]{Paul2002}), $\mathcal{E}_\Phi$ extends to a completely positive operator $\mathcal{E}$ in $\mathcal{L} (\mathbb{H}),$ i.e., a quantum operation in $\mathbb{H}$ (note that $\mathcal{E}$ is not necessarily unique). By the construction, $\mathcal{E}$ satisfies the required condition.


\end{document}